\documentclass[10pt,twocolumn,letterpaper]{article}

\usepackage{iccv}
\usepackage{times}
\usepackage{epsfig}
\usepackage{graphicx}
\usepackage{amsmath}
\usepackage{amssymb}

\usepackage{booktabs}
\usepackage{makecell}
\usepackage{multirow}

\usepackage{bbding}
\usepackage{pifont}
\usepackage{wasysym}
\usepackage{utfsym}
\usepackage{fontawesome}

\usepackage{cuted}
\usepackage{capt-of}
\usepackage[breaklinks=true,bookmarks=false]{hyperref}

\iccvfinalcopy % *** Uncomment this line for the final submission

\def\iccvPaperID{5868} % *** Enter the ICCV Paper ID here

\ificcvfinal\fi

\begin{document}

%%%%%%%%% TITLE
% \title{EEG-SD: Generating High-Quality Images from EEG Signals using Stable Diffusion}
\title{DreamDiffusion: Generating High-Quality Images from Brain EEG Signals}

% \author{First Author\\
% Institution1\\
% Institution1 address\\
% {\tt\small firstauthor@i1.org}
% % For a paper whose authors are all at the same institution,
% % omit the following lines up until the closing ``}''.
% % Additional authors and addresses can be added with ``\and'',
% % just like the second author.
% % To save space, use either the email address or home page, not both
% \and
% Second Author\\
% Institution2\\
% First line of institution2 address\\
% {\tt\small secondauthor@i2.org}
% }

\author{%
  Yunpeng Bai$^{1}$, Xintao Wang$^{2}$, Yan-Pei Cao$^{2}$, Yixiao Ge$^{2}$,  Chun Yuan$^{1, 3}$, Ying Shan$^{2}$ \\[0.5em]
  $^{1}$ Tsinghua Shenzhen International Graduate School, \\$^{2}$Tencent AI Lab, $^{3}$Peng Cheng Laboratory \\[0.3em]
%   {\small \texttt{\{chenh, bohe, hywang66,  yxren, abhinav\}@umd.edu, sernamlim@fb.com} }
}
\maketitle
% Remove page # from the first page of camera-ready.
% \ificcvfinal\thispagestyle{empty}\fi

\begin{strip}
\centering
\includegraphics[width=\textwidth]{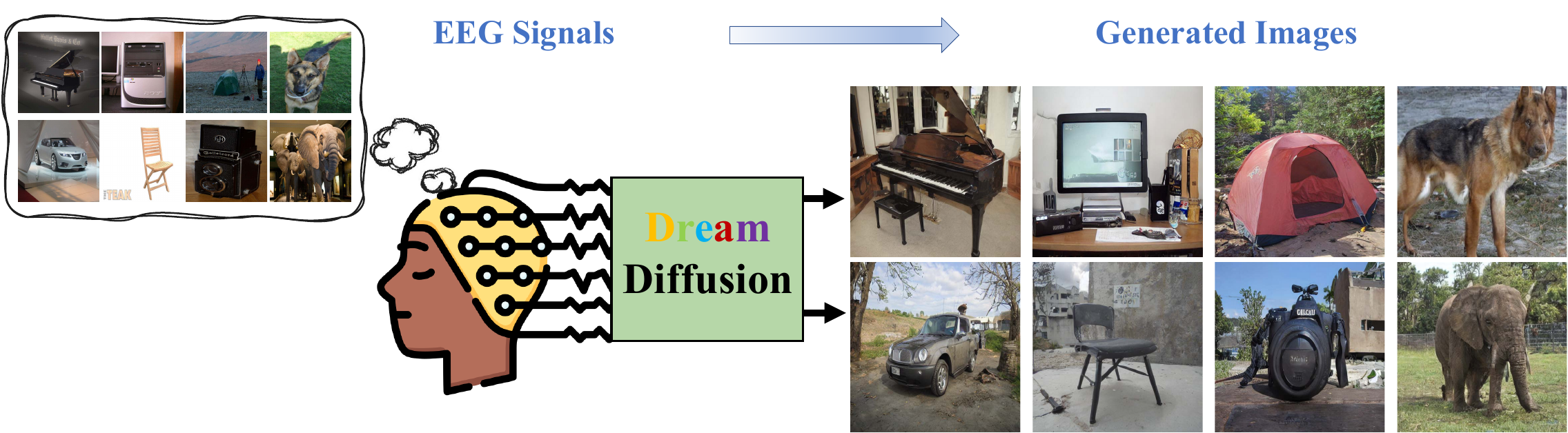}
\captionof{figure}{Our proposed DreamDiffusion is capable of generating high-quality images directly from brain electroencephalogram (EEG) signals, without the need to translate thoughts into text.}
\label{fig:teaser}
\end{strip}

%%%%%%%%% ABSTRACT
\begin{abstract}
This paper introduces DreamDiffusion, a novel method for generating high-quality images directly from brain electroencephalogram (EEG) signals, without the need to translate thoughts into text. DreamDiffusion leverages pre-trained text-to-image models and employs temporal masked signal modeling to pre-train the EEG encoder for effective and robust EEG representations. Additionally, the method further leverages the CLIP image encoder to provide extra supervision to better align EEG, text, and image embeddings with limited EEG-image pairs. Overall, the proposed method overcomes the challenges of using EEG signals for image generation, such as noise, limited information, and individual differences, and achieves promising results. Quantitative and qualitative results demonstrate the effectiveness of the proposed method as a significant step towards portable and low-cost ``thoughts-to-image'', with potential applications in neuroscience and computer vision. The code is available here \url{https://github.com/bbaaii/DreamDiffusion}.
\end{abstract}

%%%%%%%%% BODY TEXT
\section{Introduction}

Image generation~\cite{goodfellow2020generative,karras2019style,brock2018large} has made great strides in recent years, especially after breakthroughs in text-to-image generation~\cite{ramesh2021zero,ding2022cogview2,ramesh2022hierarchical,saharia2022photorealistic,bai2023textir}. The recent text-to-image generation not only dramatically improves the quality of generated images, but also enables the creation of people's ideas into exquisite paintings and artworks controlled by text.
We are very curious whether we could control image creation directly from brain activities (such as electroencephalogram (EEG) recordings), without translating our thoughts into text before creation.
This kind of ``thoughts-to-images" has broad prospects and could broaden people's imagination. 
For example, it can greatly improve the efficiency of artistic creation and help capture those fleeting inspirations. It also has the potential to help us visualize our dreams at night, (which inspires the name DreamDiffusion). Moreover, it may even aid in psychotherapy, having the potential to help children with autism and those with language disabilities.

Some recent works, such as MinD-Vis~\cite{chen2022seeing} and~\cite{takagi2022high}, attempt to reconstruct visual information based on fMRI (functional Magnetic Resonance Imaging) signals, which is another way to measure brain activities. They have demonstrated the feasibility of \textit{reconstructing} high-quality results from brain activities. 
However, they are still far away from our goal of using brain signals to create conveniently and efficiently. 
%It is impractical to use fMRI signals in the artistic generation. 
1) Since fMRI equipment is not portable and needs to be operated by professionals, it is difficult to capture fMRI signals.
2) The cost of fMRI acquisition is high. 
 They greatly hinder the widespread use of this method in the practical artistic generation.
In contrast, EEG (electroencephalogram) is a non-invasive and low-cost method of recording electrical activity in the brain.
Portable commercial products are now available for the convenient acquisition of EEG signals, showing great potential for future art generation.
% However, generating images from EEG is more challenging than fMRI due to the low spatial resolution of EEG and the limited information it contains.

In this work, we aim to leverage the powerful generative capabilities of pre-trained text-to-image models (\ie, Stable Diffusion~\cite{rombach2022high}) to generate high-quality images directly from brain EEG signals.
%supplement more relevant information. 
%In this way, we are able to generate detailed images from information-limited EEG signals. 
However, this is non-trivial and has two challenges. 
\textbf{1)} EEG signals are captured non-invasively and thus are inherently noisy. In addition, EEG data are limited and individual differences cannot be ignored. \textit{How to obtain effective and robust semantic representations from EEG signals with so many constraints?}
\textbf{2)} Thanks to the use of CLIP~\cite{radford2021learning} and the training on a large number of text-image pairs, the text and image spaces in Stable Diffusion are well aligned. 
However, the EEG signal has its own characteristics, and its space is quite different from that of text and image. \textit{How to align EEG, text and image spaces with limited and noisy EEG-image pairs?}

To address the first challenge, we propose to train EEG representations using large amounts of EEG data instead of only rare EEG-image pairs. 
Specifically, we adopt masked signal modeling to predict the missing tokens based on contextual cues. Different from MAE~\cite{he2022masked} and MinD-Vis~\cite{chen2022seeing}, which treat inputs as two-dimensional images and mask the \textit{spatial information}, we consider the temporal characteristics of EEG signals, and dig deep into the semantics behind temporal changes in people's brains. 
We randomly mask a proportion of tokens and then reconstruct those masked ones \textit{in the time domain}. In this way, the pre-trained encoder learns a deep understanding of EEG data across different people and various brain activities. 

As for the second challenge, previous methods~\cite{takagi2022high, chen2022seeing} usually directly fine-tune Stable Diffusion (SD) models using a small number of noisy data pairs.
However, it is difficult to learn accurate alignment between brain signals (\eg, EEG and fMRI) and text spaces by end-to-end fine-tuning SD only using the final image reconstruction loss.
We thus propose to employ additional CLIP~\cite{radford2021learning} supervision to assist in the alignment of EEG, text, and image spaces. 
Specifically, SD itself uses \textit{CLIP's text encoder} to generate text embeddings, which are quite different from the masked pre-trained EEG embeddings in the previous stage. 
We leverage \textit{CLIP's image encoder} to extract rich image embeddings that align well with CLIP text embeddings. Those CLIP image embeddings are then used to further optimize EEG embedding representations. 
Therefore, the refined EEG feature embeddings can be well aligned with the CLIP image and text embeddings, and are more suitable for SD image generation, which in turn improves the quality of generated images. 

Equipped with the above two delicate designs, our proposed method, namely, DreamDiffusion, can generate high-quality and realistic images from EEG signals.
Our contributions can be summarized as follows.
\textbf{1)} We propose DreamDiffusion, which leverages the powerful pre-trained text-to-image diffusion models to generate realistic images from EEG signals only. It is a further step towards portable and low-cost ``thoughts-to-images''.
\textbf{2)} A temporal masked signal modeling is employed to pre-train EEG encoder for effective and robust EEG representations.
\textbf{3)} We further leverage the CLIP image encoder to provide extra supervision to better align the EEG, text, and image embeddings with limited EEG-image pairs.
\textbf{4)} Quantitative and qualitative results have shown the effectiveness of our DreamDiffusion.

\section{Related Works}

\subsection{Generating images from brain activity}

The use of brain signals, including fMRI and EEG, to generate images has been an active area of research. For the use of fMRI, traditional methods rely on fMRI-image paired data to train the model to predict image features from fMRI. These image features will be fed into GANs~\cite{DBLP:journals/ficn/ShenDMHK19} for stimulus reconstruction during testing. However, recent studies~\cite{bird2014categorical} have proposed unsupervised approaches, such as a reconfigurable autoencoder design, to learn from unpaired fMRI and images, and utilize regression models~\cite{mozafari2020reconstructing,ozcelik2022reconstruction} to extract a latent fMRI representation that can be used to fine-tune a pre-trained conditional BigGAN~\cite{DBLP:conf/iclr/BrockDS19} for decoding. The recent work MinD-Vis~\cite{chen_2022_arXiv} integrates SC-MBM and DC-LDM to generate more plausible images with better-preserved semantic information.

Similarly, generating images from EEG signals has also been explored using deep learning techniques.  Brain2image~\cite{DBLP:conf/mm/KavasidisPSGS17} have developed using LSTM and generative methods to learn a more compact representation of EEG data for generating visual stimuli that evoke specific brain responses. ThoughtViz~\cite{DBLP:conf/mm/TirupatturRSS18} takes encoded EEG signals as input to generate corresponding images, even with limited training data. \cite{davis2022brain} uses EEG as a supervision signal for learning semantic feature representations and achieving comparable performance to semantic image editing. Overall, these approaches demonstrate the potential of using brain signals to generate images and advance the field of brain-computer interfaces.

%写几部分
% 从大脑活动中获得图像的工作
% 预训练模型
% 扩散模型 特别是Stable Diffusion

\begin{figure*}[t]
    \centering
    \includegraphics[width=\linewidth]
    {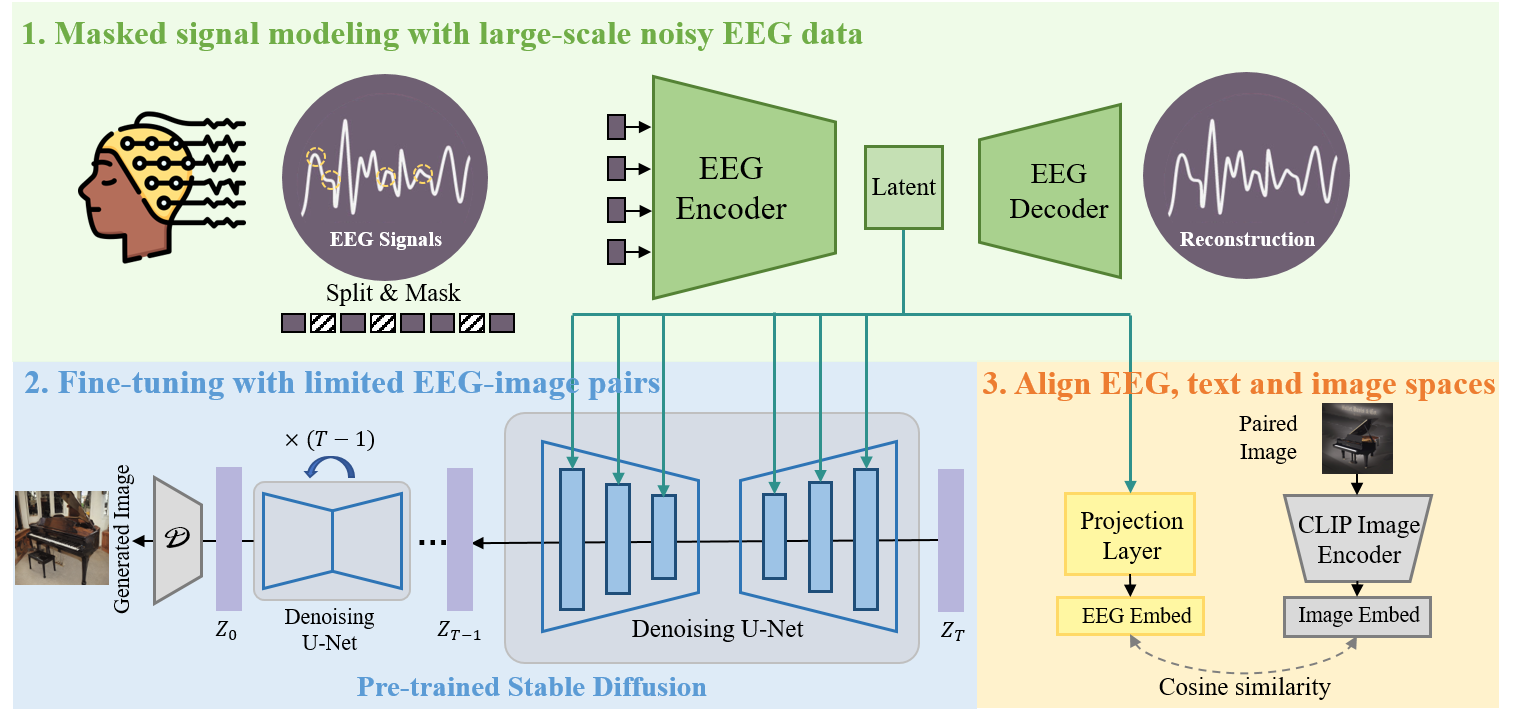}
    \caption{ \textbf{Overview of  DreamDiffusion}. Our method comprises three main components: 1) masked signal pre-training for an effective and robust EEG encoder, 2) fine-tuning with limited EEG-image pairs with pre-trained Stable Diffusion, and 3) aligning the EEG, text, and image spaces using CLIP encoders.  }
    \label{fig:pipeline}   
    % \vspace{-0.1in}
\end{figure*}

\subsection{Model pre-training}

Pre-training models have become increasingly popular in the field of computer vision, with various self-supervised learning approaches focusing on different pretext tasks~\cite{doersch2015unsupervised, wang2015unsupervised, noroozi2016unsupervised}. These methods often utilize pretext tasks such as contrastive learning~\cite{becker1992self,hadsell2006dimensionality}, which models image similarity and dissimilarity, or autoencoding~\cite{ brown2020language}, which recovers the original data from a masked portion. In particular, masked signal modeling (MSM) has been successful in learning useful context knowledge for downstream tasks by recovering the original data from a high mask ratio for visual signals~\cite{he2022masked, xie2022simmim} and a low mask ratio for natural languages \cite{devlin2018bert, radford2019language}. Another recent approach, CLIP~\cite{radford2021learning}, builds a multi-modal embedding space by pre-training on 400 million text-image pairs collected from various sources on the Internet. The learned representations by CLIP are extremely powerful, enabling state-of-the-art zero-shot image classification on multiple datasets, and providing a method to estimate the semantic similarity between text and images.

\subsection{Diffusion models}

Diffusion models have become increasingly popular as generative models for producing high-quality content~\cite{sohl2015deep}. The basic form of diffusion models is a probabilistic model defined by a bi-directional Markov Chain of states~\cite{ho2020denoising}. These models~\cite{dhariwal2021diffusion,ho2020denoising,ronneberger2015u,DBLP:conf/iclr/0011SKKEP21} exhibit strong generative power due to their natural fit with the inductive biases of image-like data. The best synthesis quality is typically achieved when using a reweighted objective during training~\cite{ho2020denoising}, allowing for a trade-off between image quality and compression capabilities. However, evaluating and optimizing these models in pixel space is computationally expensive and time-consuming~\cite{DBLP:journals/corr/abs-2106-00132,DBLP:journals/corr/abs-2104-02600,DBLP:conf/iclr/SongME21,DBLP:journals/jmlr/HoSCFNS22,DBLP:conf/nips/VahdatKK21}.

To address these challenges, some diffusion models work on a compressed latent space of lower dimensionality, such as the proposed LDMs~\cite{rombach2022high}. By compressing images into lower-dimensional latent features using a Vector Quantization (VQ)~\cite{esser2021taming} regularized autoencoder and then reconstructing them using the same latent space features, the LDM reduces computational costs while maintaining synthesis quality. Additionally, a UNet-based denoising model with attention modules offers the flexibility to condition image generation through key/value/query vectors during Markov Chain transitions. This approach has several advantages, including reduced computational costs and  better quality of image synthesis.
%-------------------------------------------------------------------------

% Our framework consists of an encoder and a generator. The encoder is used to extract features of the degraded images for fusion with the generated features. The generator is used to generate the restoration results. During training, we use the ground truth as the condition. With the help of CLIP's text-image shared feature space, we can use text as a condition to obtain results that match the description during inference.

\section{Proposed Method}
Our method comprises three main components: 1) masked signal pre-training for an effective and robust EEG encoder, 2) fine-tuning with limited EEG-image pairs with pre-trained Stable Diffusion, and 3) aligning the EEG, text, and image spaces using CLIP encoders. 
Firstly, we leverage masked signal modeling with lots of noisy EEG data to train an EEG encoder to extract contextual knowledge. The resulting EEG encoder is then employed to provide conditional features for Stable Diffusion via the cross-attention mechanism. 
In order to enhance the compatibility of EEG features with Stable Diffusion, we further align the EEG, text, and image embedding spaces by reducing the distance between EEG embeddings and CLIP image embeddings during fine-tuning. 
As a result, we can obtain DreamDiffusion, which is capable of generating high-quality images from EEG signals only.

%先写预训练
%再写微调的部分 
%有一部分是写训练目标

\begin{figure}[t]
    \centering
    \includegraphics[width=\linewidth]
    {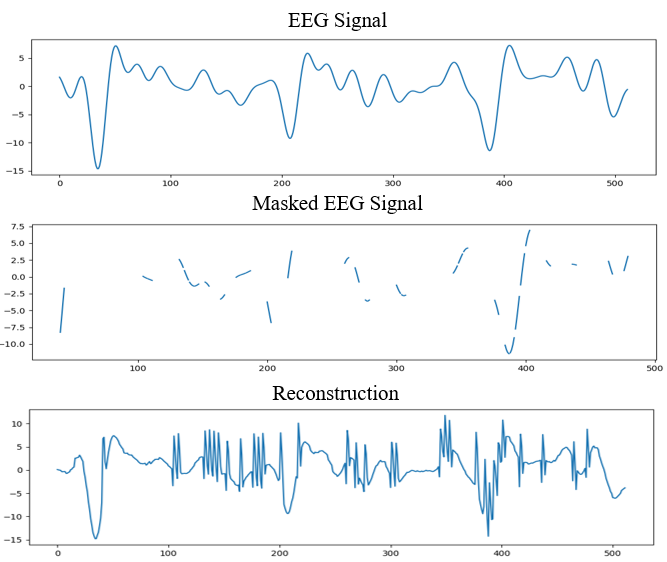}
    \caption{Masked signals modeling with large-scale noisy EEG data. We visualize the reconstruction results of one channel from the EEG data. We can observe that the overall trend is accurate, but the details are influenced by the dataset, as the EEG signals in these datasets are relatively noisy.}
    \label{fig:msm}   
    % \vspace{-0.1in}
\end{figure}

\subsection{Masked signal pre-training for effective and robust EEG representations}
EEG (Electroencephalogram) data is a recording of electrical activity generated by the human brain, measured using electrodes placed on the scalp. It is a non-invasive and low-cost method of measuring brain activity. EEG data has several characteristics. Firstly, the data is two-dimensional, with one dimension representing the channels or electrodes placed on the scalp, and the other dimension representing time. The temporal resolution of EEG is high, meaning that it can capture rapid changes in brain activity that occur on the order of milliseconds. However, the spatial resolution of EEG is low, meaning that it is difficult to precisely localize the source of the activity within the brain. Secondly, EEG signals are highly variable, influenced by factors such as age, sleep, and cognitive state. Finally, EEG data is often noisy, and requires careful processing and analysis to extract meaningful information. 

Due to the inherent variability and noise in EEG data, conventional modeling methods often struggle to extract meaningful information from EEG signals. Consequently, adopting masked signal modeling techniques, which have been proven effective in capturing contextual information from noisy and variable data~\cite{he2022masked,chen2022seeing}, represents a promising avenue for deriving meaningful contextual knowledge from large-scale noisy EEG data. 
Different from MAE~\cite{he2022masked} and MinD-Vis~\cite{chen2022seeing}, which treat inputs as two-dimensional images and mask the \textit{spatial information}, we consider the temporal characteristics of EEG signals, and dig deep into the semantics behind temporal changes in people's brains. 

Given the high temporal resolution of EEG signals, we first divide them into tokens in the time domain, and randomly mask a certain percentage of tokens. Subsequently, these tokens will be transformed into embeddings by using a one-dimensional convolutional layer.
Then, we use an asymmetric architecture such as MAE \cite{he2022masked} to predict the missing tokens based on contextual cues from the surrounding tokens. Through reconstructing the masked signals, the pre-trained EEG encoder learns a deep understanding of EEG data across different people and various brain activities.

\begin{figure*}[t]
    \centering
    \includegraphics[width=\linewidth]
    {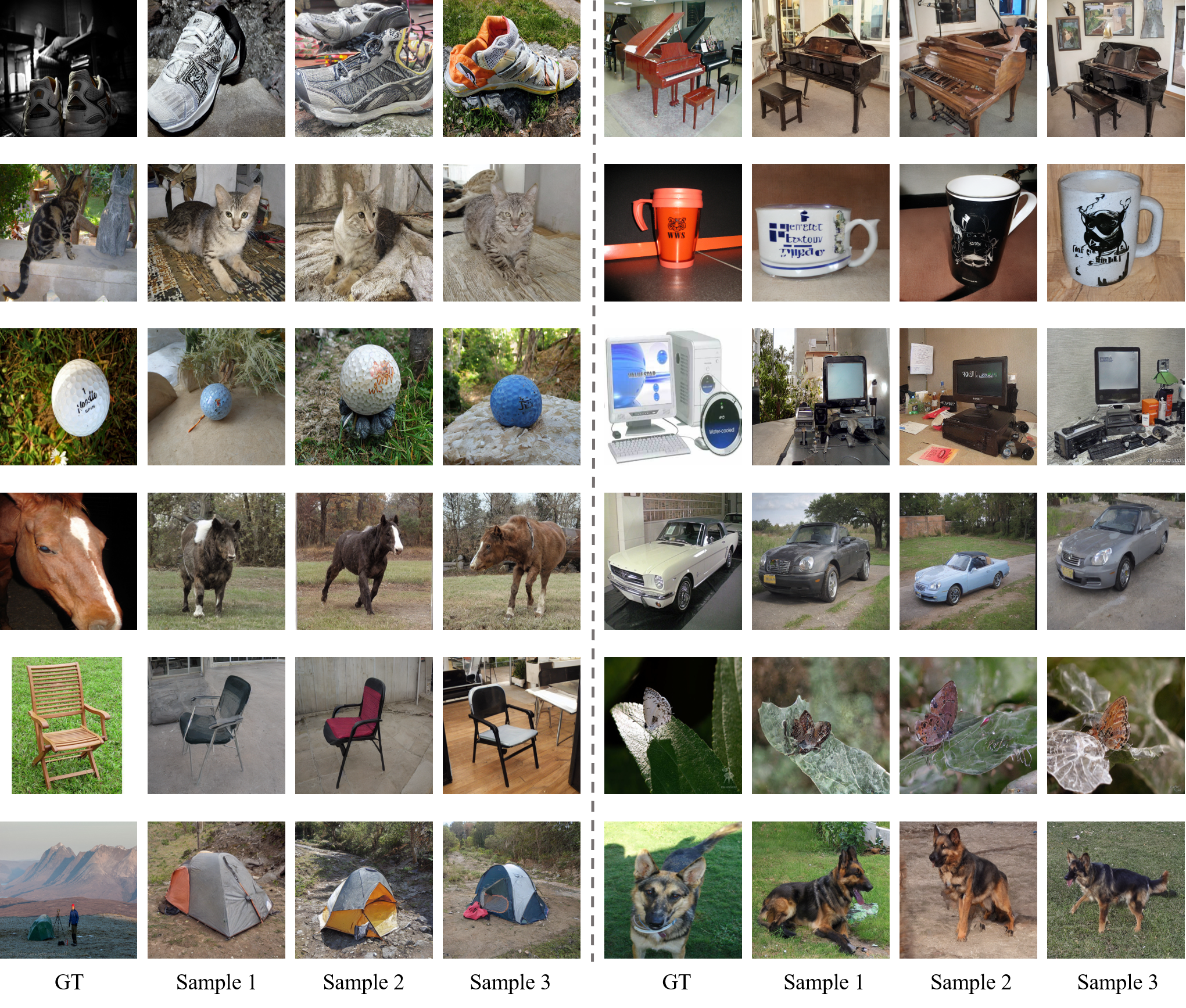}
    \caption{ \textbf{Main results.} The images on the left depict paired image data, while the three images on the right represent the sampling results. It can be observed that our model generates images of high quality from the EEG data, and these images match the EEG data accurately.}
    \label{fig:results}   
    % \vspace{-0.1in}
\end{figure*}

\subsection{Fine-tuning with Stable Diffusion on limited EEG-image pairs}
After obtaining an effective representation of EEG signals from masked signal pre-training, we utilize it to generate images by leveraging a pre-trained Stable Diffusion (SD) model. 
Stable Diffusion involves gradually denoising a normally distributed variable to learn a data distribution. 
%, using a reweighted variant of the variational lower bound. 
SD is augmented with a cross-attention mechanism for more flexible conditional image generation and the most common condition is the text prompt. 
Stable Diffusion has shown great generative power in  generating high-quality images from various types of signals, such as labels, text, and semantic maps.

Stable Diffusion operates on the latent space.  Given an image $x$ in pixel space, $x$ is encoded by a VQ encoder $\mathcal{E}(\cdot)$ to obtain the corresponding latent $z = \mathcal{E}(x)$. 
Conditional signals are introduced by the cross-attention mechanism in the UNet. 
This cross-attention can also incorporate conditional information from the EEG data. 
Specifically, the output of EEG encoder $y$ is further projected with a projector $\tau_\theta$ into an embedding $\tau_\theta(y) \in \mathbb{R}^{M \times d_\tau}$.
Then, this EEG representation is incorporated into U-Net by a cross-attention layer implementing $\operatorname{Attention}(Q, K, V)=\operatorname{softmax}\left(\frac{Q K^T}{\sqrt{d}}\right) \cdot V$.

\begin{equation}
Q=W_Q^{(i)} \cdot \varphi_i\left(z_t\right), K=W_K^{(i)} \cdot \tau_\theta(y), V=W_V^{(i)} \cdot \tau_\theta(y),
\end{equation}

where $\varphi_i\left(z_t\right) \in \mathbb{R}^{N \times d_e^i}$ denotes intermediate values of the U-Net.
$W_V^{(i)} \in \mathbb{R}^{d \times d_\epsilon^i}, W_Q^{(i)} \in \mathbb{R}^{d \times d_\tau}\ \text{and}\ W_K^{(i)} \in \mathbb{R}^{d \times d_\tau}$ are projection matrices with learnable parameters.
%The generating process is referred to as the reverse diffusion process $q\left(x_{t-1} \mid x_t, y\right)$

During the fine-tuning process, we optimize the EEG encoder and cross-attention heads of the U-Net together. We keep the remaining parts of Stable Diffusion fixed. We use the following SD loss function for fine-tuning.

\begin{equation}
L_{SD}=\mathbb{E}_{x, \epsilon \sim \mathcal{N}(0,1), t}\left[\left\|\epsilon-\epsilon_\theta\left(x_t, t, \tau_\theta(y)\right)\right\|_2^2\right],
\end{equation}
where $\epsilon_\theta$ is the denoising function implemented as UNet.

\begin{figure*}[t]
    \centering
    \includegraphics[width=\linewidth]
    {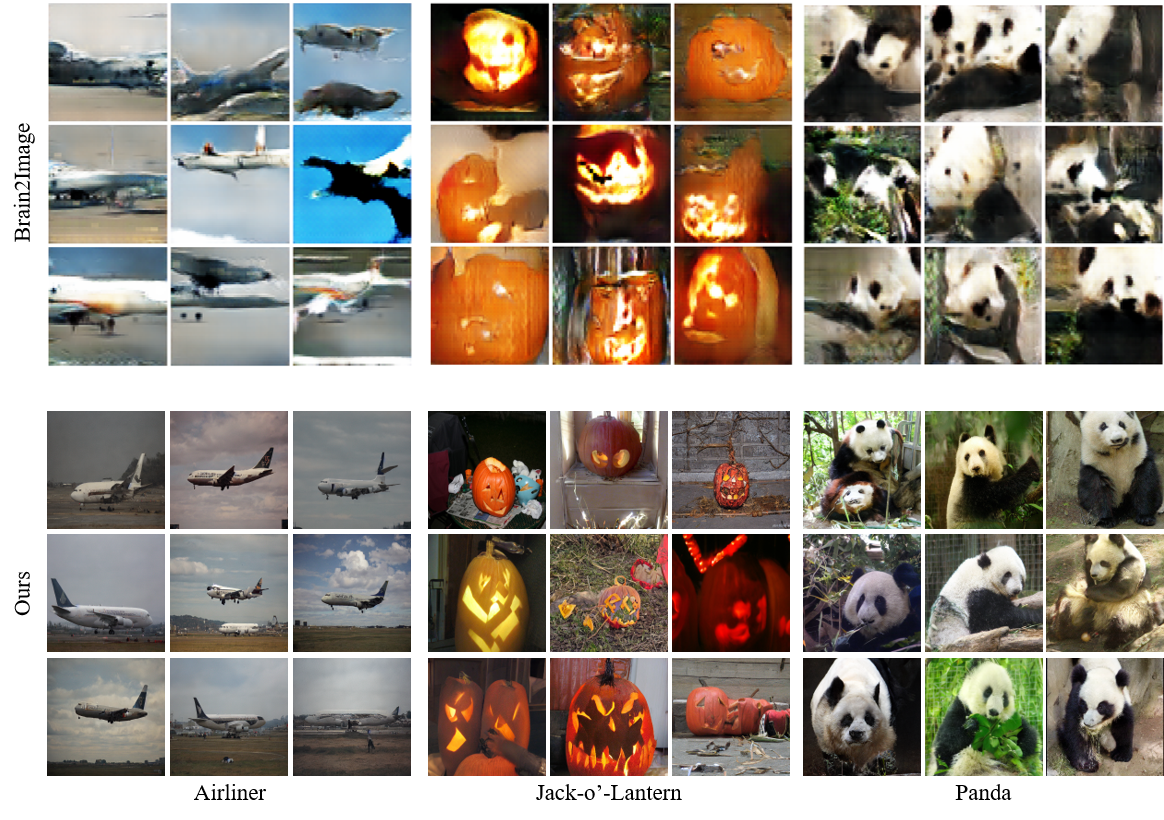}
    \caption{ \textbf{Comparison with Brain2Image.} The quality of the generated images produced by DreamDiffusion is significantly higher than those generated by Brain2Image.}
    \label{fig:com}   
    % \vspace{-0.1in}
\end{figure*}

\subsection{Aligning the EEG, text, and image spaces with CLIP encoders}
Next, we will fine-tune the EEG representation obtained from pre-training to make it more suitable for generating images. 
The pre-trained Stable Diffusion model is specifically trained for text-to-image generation; however, the EEG signal has its own characteristics, and its latent space is quite different from that of text and image.
Therefore, directly fine-tuning the Stable Diffusion model end-to-end using limited EEG-image paired data is unlikely to accurately align the EEG features with existing text embedding in pre-trained SD.

Thanks to the use of CLIP~\cite{radford2021learning} and the training on a large number of text-
image pairs, the text and image spaces in Stable Diffusion
are well aligned. 
Therefore, we propose to employ additional CLIP~\cite{radford2021learning} supervision to assist in the alignment of EEG, text, and image space.
Specifically, the EEG features obtained from the pre-trained encoder are transformed into embeddings with the same dimension as those of CLIP through a projection layer. We then use a loss function to minimize the distance between the EEG embeddings and the image embeddings obtained from the CLIP image encoder.
The CLIP model is fixed during the fine-tuning process.
The loss function is defined as follows:

\begin{equation}
    \mathcal{L}_{clip} = 1 - \frac{E_I(I)\cdot h(\tau_\theta(y))}{|E_I(I)||h(\tau_\theta(y))|},
\end{equation}
where $h$ is a projection layer and $E_I$ is the CLIP image encoder. This loss function can encourage the EEG features to become more closely aligned with the image and thus more similar to text features.
In this way, we can align the EEG signal, text and image in one unified space. 
The optimized EEG embedding representation is more suitable for SD image
generation, which in turn improves the quality of generated
images.

\begin{table*}
    % \footnotesize
    \small
    \centering 
    
\begin{tabular}{ccccccc}
\hline Model &  MSM Pretraining& CLIP Finetuning & Mask Ratio & E + A & Params & Acc (\%) \\
\hline Full & \Checkmark & \Checkmark & $\mathbf{0 . 7 5}$ & E + A & $\mathbf{297 M}$ & $\mathbf{45.8} $ \\
\hline \hline 1 & \XSolidBrush & \XSolidBrush & - & E + A & $\mathbf{297 M}$ & $4.2 $ \\

2 & \XSolidBrush & \XSolidBrush & - & E + A & $\mathbf{18.3 M}$ & $3.7 $ \\
3 & \XSolidBrush & \Checkmark & - & E + A & $\mathbf{297 M}$ & $32.3 $ \\
4 & \XSolidBrush & \Checkmark & - & E + A & $\mathbf{18.3 M}$ & $24.5 $ \\
\hline 
5 & \Checkmark & \Checkmark & 0.25 & E + A & $\mathbf{297 M}$ & $19.7 $ \\
6 & \Checkmark & \Checkmark & 0.5 & E + A & $\mathbf{297 M}$ & $38.3 $ \\
7 & \Checkmark & \Checkmark & 0.85 & E + A & $\mathbf{297 M}$ & $33.4 $ \\

\hline 
8 & \Checkmark & \Checkmark & 0.75 & E + A & $\mathbf{458 M}$ & $38.5$ \\
9 & \Checkmark & \Checkmark & 0.75 & E + A & $\mathbf{162 M}$ & $36.6$ \\
10 & \Checkmark & \Checkmark & 0.75 & E + A & $\mathbf{74 M}$ & $29.8 $ \\
11 & \Checkmark & \Checkmark & 0.75 & E + A & $\mathbf{18.3 M}$ & $28.7$ \\
\hline 
12 & \Checkmark & \Checkmark & 0.75 & E only & $\mathbf{297 M}$ & $22.4$ \\
\hline 
13 & \Checkmark & \XSolidBrush & 0.75 & E + A & $\mathbf{297 M}$ & $28.3 $ \\
14 & \Checkmark & \XSolidBrush & 0.75 & A only & $\mathbf{297 M}$ & $20.9 $ \\
\hline
\end{tabular}

\vspace{0.5em}
\caption{\textbf{Quantitative results of ablation studies.} E and A represent fine-tuning of the encoder and cross-attention heads, respectively.}

    \label{tab:ablation1}
\end{table*}

%-------------------------------------------------------------------------

\section{Experiments and Analyses}
\subsection{Implementation details}

\noindent\textbf{Data for EEG representation pre-training.} We have collected approximately 120,000 EEG data samples from over 400 subjects with channel ranges from 30 to 128 on the MOABB \cite{jayaram2018moabb} platform for the EEG pre-training. MOABB is a software package designed to facilitate the development of brain-computer interface (BCI) algorithms by providing a collection of publicly available EEG datasets in a common format, along with a suite of state-of-the-art algorithms. This platform enables researchers to easily validate new algorithms using automated statistical analysis, eliminating the need for time-consuming and unreliable data preprocessing. These data contain a wide variety of EEG data, including tasks such as looking at an object, motor imagery, and watching videos. Our goal is to learn universal representations from diverse EEG data, without specific requirements on the types of EEG data. Due to variations in the equipment used for data acquisition, the channel counts of these EEG data samples differ significantly. To facilitate pre-training, we have uniformly padded all the data to 128 channels by filling missing channels with replicated values. During the pre-training process, every 4 adjacent time steps are grouped into a token and each token is transformed into a 1024-dimensional embedding through a projection layer for subsequent masked signal modeling. The loss function calculates the MSE between the reconstructed and original EEG signals. The loss is only computed on masked patches. The reconstruction is performed on the entire set of 128 channels as a whole, rather than on a per-channel basis. The decoder is discarded after pretraining.

% The channels of pretraining data ranges from 30 to 128 channels, rather than around 30. To facilitate training with all data together, we pad the data that has fewer than 128 channels. The reconstruction is performed on the entire set of 128 channels as a whole, rather than on a per-channel basis. These data contain a wide variety of EEG data, including tasks such as looking at an object, motor imagery, and watching videos. Our goal is to learn universal representations from diverse EEG data, without specific requirements on the types of EEG data.
% \textbf{Training methods:}
%  The objective of pretraining is to reconstruct the input signal. The loss function calculates the MSE between the reconstructed and original EEG signals. The loss is only computed on masked patches.
% \textbf{EEG preprocessing:}
%  All EEG signals are filtered within the frequency range of 5-95 Hz before pretraining. Subsequently, the signals are truncated to a common length of 512.
% \textbf{Others:}
%  Epoch for pretraining is 500 (L676, main paper). Temporal token has defined in L635 in the main paper.
%  Thanks for all suggestions, we will add those details.

\noindent\textbf{Paired EEG-image data.} We adopt the ImageNet-EEG \cite{DBLP:conf/mm/KavasidisPSGS17} dataset for our ``thoughts-to-image'' experiments, which is a collection of EEG recordings obtained from 6 subjects while they were shown 2000 images belonging to 40 different categories of objects from the ImageNet dataset. Each category consisted of 50 images, and each image was presented for 0.5 seconds, followed by a 10-second pause for every 50 images. The EEG data were recorded using a 128-channel Brainvision EEG system, resulting in a total of 12000 128-channel EEG sequences. The dataset includes images of various objects, such as animals (dogs, cats, elephants, etc.), vehicles (airliners, bikes, cars, etc.), and everyday objects (computers, chairs, mugs, etc.). More details can be found in the related reference \cite{DBLP:conf/mm/KavasidisPSGS17}.

\noindent\textbf{Other implementation details.} We use version 1.5 of Stable Diffusion for image generation. The mask ratio for EEG signals is set to $75\%$. All EEG signals are filtered within the frequency range of 5-95 Hz before pretraining. Subsequently, the signals are truncated to a common length of 512. The encoder is pre-trained for 500 epochs and finetuned with Stable Diffusion for another 300. The pre-training model for EEG is similar to ViT-Large in \cite{dosovitskiy2020image}. The training and testing were conducted on the same subject, and all results presented in the paper were generated using data from Subject 4.

% MOABB's consistent interface for machine learning methods streamlines the development process and promotes reproducibility in BCI research. Overall, MOABB serves as a valuable tool for researchers seeking to advance the field of brain-computer interfaces.

% \subsection{Qualitative comparison}

\subsection{Comparison with Brain2Image~\cite{DBLP:conf/mm/KavasidisPSGS17}}

In this section, we present a comparison of our proposed approach with Brain2Image~\cite{DBLP:conf/mm/KavasidisPSGS17}, a recent work that employs conventional generative models, i.e., variational autoencoders (VAE) and generative adversarial networks (GAN), to achieve EEG-to-images. Brain2Image, however, presents results for only a few categories and does not provide a reference implementation. In light of this, we conducted a qualitative comparison of the results on a few categories (namely, Airliner, Jack-o-Lantern, and Panda) that were showcased in the Brain2Image paper. To ensure a fair comparison, we followed the same subjective evaluation strategy as outlined by Brain2Image and presented generated instances of different methods in Figure~\ref{fig:com}. The top rows depict the results generated by Brain2Image, whereas the bottom rows were generated by our proposed method, DreamDiffusion. We observed that the quality of the generated images produced by DreamDiffusion is significantly higher than those generated by Brain2Image, thus validating the efficacy of our proposed method.

\begin{figure*}[!t]
    \centering
    \includegraphics[width=\linewidth]
    {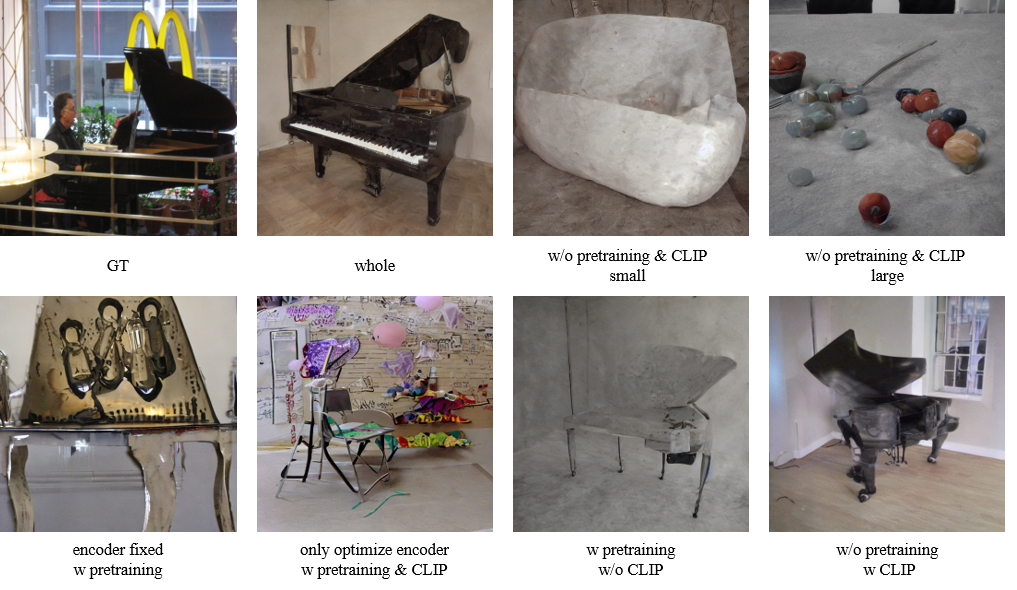}
    \caption{ \textbf{Qualitative results of ablation studies.} }
    \label{fig:ablation}   
    % \vspace{-0.1in}
\end{figure*}

\subsection{Ablation studies}

In this section, we conduct several ablation studies on the proposed framework using various cases. We evaluate the effectiveness of different methods by employing a 50-way top-1 accuracy classification task. We use a pre-trained ImageNet1K classifier \cite{dosovitskiy2020image} to determine the semantic correctness of the generated images. Both the ground-truth and generated images will be inputted into the classifier. Then, we will verify whether the top-1 classification of the generated image matches the ground-truth classification in 50 selected classes. A generated image will be deemed correct as long as the semantic classification results of the generated image and the ground-truth are consistent.

\noindent\textbf{Role of pre-training.}
To demonstrate the effectiveness of the pretraining with large-scale EEG data, we conduct a validation by training several models with untrained encoders. One of the models is identical to the full model, while the other model has a shallow EEG encoding layer with only two layers to avoid overfitting the data. During the training process, the two models were trained with and without clip supervision, and the results are shown in Table \ref{tab:ablation1}, Model 1-4. It can be observed that the accuracy of the model without pre-training decreased.

\noindent\textbf{Mask ratios.} We investigate to determine the optimal mask ratio for MSM pretraining with EEG data. As shown in Model 5-7 of Table \ref{tab:ablation1}, excessively high or low mask ratios can have a detrimental effect on the model's performance. The highest overall accuracy was achieved at a mask ratio of 0.75. This finding is significant as it suggests that, unlike natural language processing where low mask ratios are commonly used, a high mask ratio is also a preferable option when performing MSM on EEG.

% We applied a similar approach to validate the appropriate mask ratio for pretraining with EEG data. In Table 1, Model 10-14, we demonstrate that a high mask ratio does not hinder the decoding performance at the outset, with the most exceptional overall accuracy obtained at a mask ratio of 0.75. Notably, employing a high mask ratio offers the benefit of significantly reducing memory usage since the encoder only operates on unmasked patches. This is particularly important for EEG data, as our pretraining model SC-MBM is more memory-intensive than MIM, as a result of the higher embedding-to-patch-size ratio.

\noindent\textbf{CLIP aligning.} One of the keys of our method is to align the EEG representation with the image through the CLIP encoder. To validate the effectiveness of this approach, we conducted experiments 13-14 as shown in Table \ref{tab:ablation1}. It can be observed that the performance of the model significantly decreases when CLIP supervision is not used. In fact, as shown in the bottom right corner of Figure \ref{fig:ablation}, even in the absence of pre-training, using CLIP to align EEG features can still yield reasonable results, which highlights the importance of CLIP supervision in our method.

\begin{figure}[h]
    \centering
    \includegraphics[width=\linewidth]
    {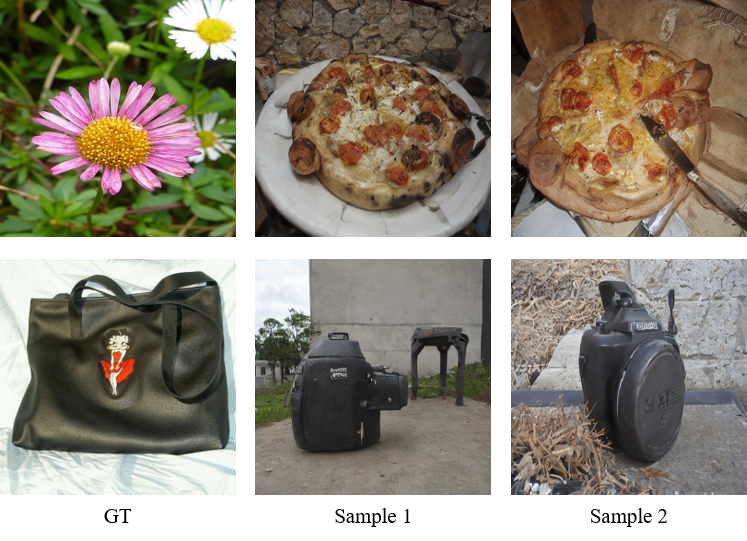}
    \caption{ \textbf{Failure cases of DreamDiffusion.} }
    \label{fig:failure}   
    % \vspace{-0.1in}
\end{figure}

\section{Conclusion}

% In this paper, we propose using EEG signals as an alternative source of brain activity to generate images. One of the main advantages of EEG signals is that they are non-invasive and can be easily obtained. However, using EEG signals for image generation presents several challenges, which we address by utilizing the knowledge learned from large EEG datasets and the powerful generative capabilities of diffusion models. We propose a new method, DreamDiffusion, that overcomes the challenges associated with generating images from EEG signals and has the potential to be used in a wide range of applications. We have designed a pre-training and fine-tuning scheme for EEG data to obtain EEG representations suitable for image generation. With the powerful image generation capabilities of stable diffusion, we can generate high-quality images from EEG signals. Overall, our method represents an important step forward in the field of image generation from brain activity.
This paper proposes a novel method, DreamDiffusion, for generating high-quality images from EEG signals, which is a non-invasive and easily obtainable source of brain activity. The proposed method addresses the challenges associated with EEG-based image generation by utilizing the knowledge learned from large EEG datasets and the powerful generative capabilities of image diffusion models. Through a pre-training and fine-tuning scheme, EEG data can be encoded to the representation suitable for image generation using Stable Diffusion. Our method represents a significant advancement in the field of image generation from brain activity.

\noindent\textbf{Limitations.} Currently, EEG data only provide coarse-grained information at the category level in experimental results. Figure~\ref{fig:failure} shows some failure cases, where some categories are mapped to other categories with similar shapes or colors. We assume this may be due to the fact that the human brain considers shape and color as two important factors when recognizing objects. Nevertheless, DreamDiffusion has the potential to be used in a wide range of applications, such as neuroscience, psychology, and human-computer interaction.

{\small
\bibliographystyle{ieee_fullname}
\bibliography{egbib}
}

\end{document}